\documentclass{article}

     \usepackage[preprint]{neurips_2020}

\usepackage[utf8]{inputenc} 
\usepackage[T1]{fontenc}    
\usepackage{hyperref}       
\usepackage{url}            
\usepackage{booktabs}       
\usepackage{amsfonts}       
\usepackage{nicefrac}       
\usepackage{microtype}      
\usepackage{xcolor}
\usepackage{bm}
\usepackage{graphicx}
\usepackage{breqn}
\usepackage{amssymb}
\usepackage{amsmath}
\usepackage{multicol}
\usepackage{natbib}

\usepackage{subfigure}

\newcommand{\todo}[1]{\textcolor{red}{\{TODO: #1\}}}
\newcommand{\comment}[1]{}

\title{Three-Player Game Training Dynamics}

\author{%
  Kenneth ~Christofferson$^*$\\
  Department of Computer Science\\
  University of Toronto\\
  \texttt{kench@cs.toronto.edu} \\
   \And
  Fernando J.~Yanez$^*$\\
  Department of Computer Science\\
  University of Toronto\\
  \texttt{fyanez@cs.toronto.edu} \\
}

\begin{document}

\maketitle
\def\thefootnote{*}
\footnotetext{Equal contribution. Listed alphabetically.}

    
\begin{abstract}
This work explores three-player game training dynamics, under what conditions three-player games converge and the equilibria the converge on. In contrast to prior work, we examine a three-player game architecture in which all players explicitly interact with each other. Prior work analyzes games in which two of three agents interact with only one other player, constituting dual two-player games. We explore three-player game training dynamics using an extended version of a simplified bilinear smooth game, called a simplified trilinear smooth game. We find that trilinear games do not converge on the Nash equilibrium in most cases, rather converging on a fixed point which is optimal for two players, but not for the third. Further, we explore how the order of the updates influences convergence. In addition to alternating and simultaneous updates, we explore a new update order—maximizer-first—which is only possible in a three-player game. We find that three-player games can converge on a Nash equilibrium using maximizer-first updates. Finally, we experiment with differing momentum values for each player in a trilinear smooth game under all three update orders and show that maximizer-first updates achieve more optimal results in a larger set of player-specific momentum value triads than other update orders.

\end{abstract}

\section{Introduction}

Generative Adversarial Networks (GANs) \citep{goodfellow2014generative} have been well explored. In their paper, Goodfellow et. al describe how an adversarial relationship between networks can be used to produce a network capable of creating samples that are very similar to a target distribution. This is achieved by training two networks, one which produces a sample and another which determines whether the sample produced is real or fake. GANs have been extensively studied and extended, becoming one of the standard generative network architectures.

Much recent work has been dedicated to better understanding two-player GANs optimization dynamics \citep{heusel2017gans, li2018limitations, mescheder2017numerics, miyato2018spectral, nie2020towards, chao2020improved}. This is unsurprising, as many famous GAN use cases rely on a two-player architecture—usually a Generator and a Discriminator. However, GANs are not limited to two-player architectures and some problems might require additional players. For example \cite{li2017triple} and \cite{vandenhende2019three} explore three-player GANs, including also a Classifier. Interestingly, \cite{li2017triple} performs a theoretical analysis of convergence for the three-player game they proposed, while \cite{vandenhende2019three} only shows experimental results from their own game. Thus, none of such papers introduce a thorough study of the nature of the game they propose, particularly on the interactions between the players—or lack thereof.

\begin{figure}[t]
    \subfigure[Three-Player GAN]{\includegraphics[width=0.32\textwidth]{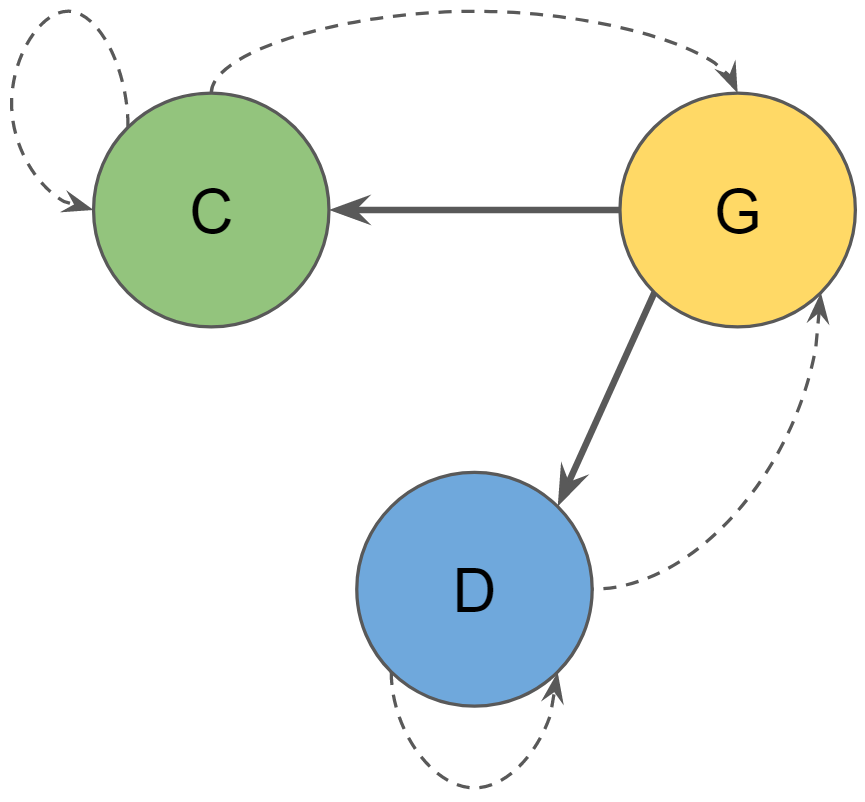}}
    \subfigure[Triple-GAN]{\includegraphics[width=0.32\textwidth]{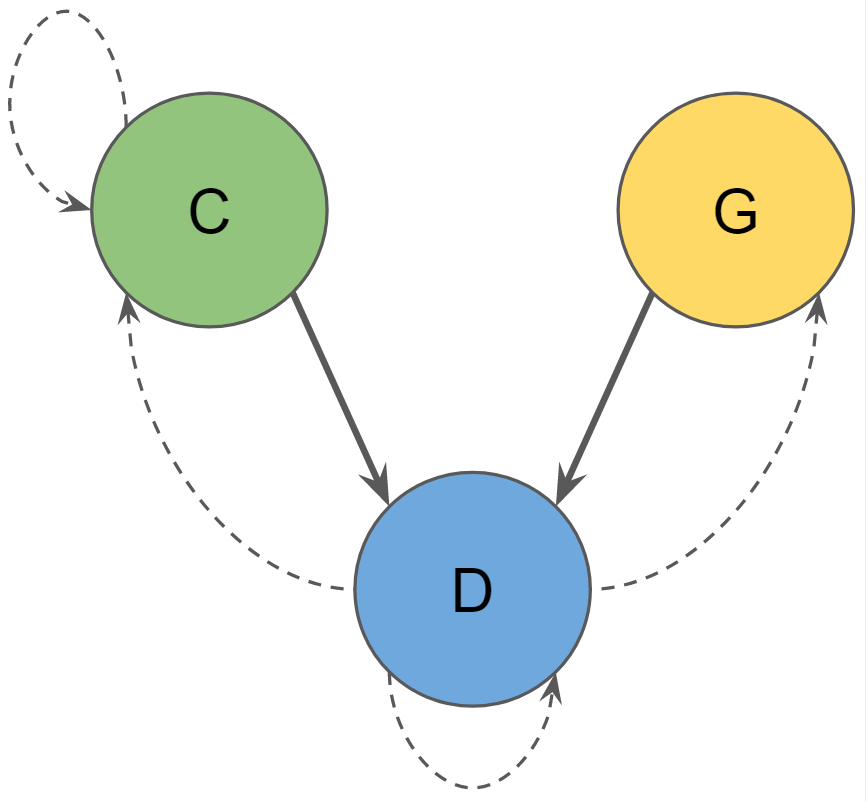}}
    \subfigure[SquidGAN]{\includegraphics[width=0.32\textwidth]{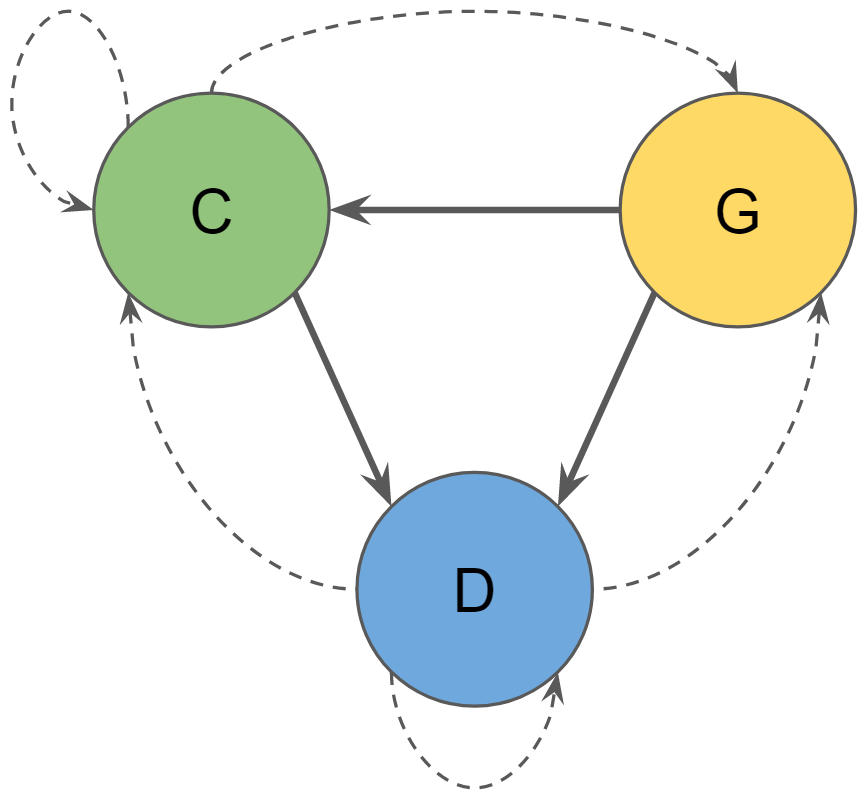}}
    \caption{Dependencies between each network in the different GAN models with three players. Each node represents a NN: C (green) is the Classifier, G (yellow) the Generator, and D (blue) the Discriminator. Each edge represent the data dependency (solid line) or the back-propagation dependency (dotted line) between networks.}
    \label{fig:intro}
\end{figure}

In this paper, we analyze the dynamics of simple three-player games and propose SquidGAN: a three-player game where all the agents interact with each other. First, show the resemblance between \cite{li2017triple}'s Triple-GAN and \cite{vandenhende2019three}'s Three-Player GAN with dual bilinear smooth games in \S \ref{sec:TPG}, showing the matrix form of the alternating update rule with momentum\footnote{Given \cite{gidel2019negative}'s conclusions on the benefits of performing alternating gradient steps with negative momentum for two-player games, we perform the main analyses in this paper using such an update rule.}, followed by the definition of a simplified trilinear smooth game. We then define SquidGAN as a min-max-max game. Lastly, we show the results of the three smooth games mentioned, as well as the performance of SquidGAN over a Gaussian classification task. Figure \ref{fig:intro} shows an illustration of the back-propagation dependencies per network, for each of the GAN models with three players so far.

\section{Related Work}

\cite{gidel2019negative} analyse the use of negative momentum in order to improve two-player games dynamics. Further, the authors found that alternating gradient descent improved the dynamics compared while the alternative, simultaneous updates. Their analysis includes both theoretical and empirical components, which we draw inspiration from. Specifically, they explore two-player GAN training dynamics using a simplified smooth game, which we extend for three players.

Two-player GAN training dynamics are well studied because they are often unstable. For example, \cite{mescheder2017numerics} analyzed two-player GAN gradient vector fields, concluding that the algorithm's they studied converged slowly or failed to converge (i.e., failed to find local Nash-equilibrium) due to two factors: 1) the presence of complex eigenvalues with a zero real component in the Jacobian of the gradient vector field, and 2) eigenvalues with large complex components. Building upon that, \cite{heusel2017gans} proposed a method to ameliorate such convergence failures by varying the learning rates applied to the Discriminator and Generator networks called Two Time Scale Update Rule (TTUL). They show that TTUL converges to the local Nash equilibrium when applied to both stochastic gradient descent and Adam optimization. Further, \cite{li2018limitations} analyzed GAN dynamics with certain behaviors like vanishing gradients, mode collapse, or diverging or oscillatory behavior. In their study the authors found that a possible cause could be the use of first order Discriminator steps. They also reach to the conclusion that a GAN with an optimal Discriminator has more chance of converging. Another direction borrowed from control theory is that of \cite{xu2020understanding} while assessing how best to stabilize Dirac GAN training. Specifically, the authors consider how closed-loop control (CLC) theory could by applied to Dirac GANs, propose a training method grounded in CLC that stabilizes Dirac GAN training dynamics without altering the desired equilibrium point.

 \cite{balduzzi2018mechanics} studied the general case of n-player differentiable games. They proposed to decompose the second-order dynamics into two components which results into Symplectic Gradient Adjustment as a new algorithm that is able to find stable fixed points in general games. The authors analyzed three important problems with Gradient Descent in these kinds of games, as well as a comparison with \cite{mescheder2017numerics}'s results mentioned bellow.

 Closest to this work, \cite{li2017triple} and \cite{vandenhende2019three} explore three-player GANs including a Generator a Discriminator and a Classifier, albeit for different motivations. \cite{vandenhende2019three} use a three-player GAN to generate `hard to classify' data which is used to update the Classifier included in the GAN. They assessed the difficulty of classification based on the classification loss for each generated image. While the assumption that classification loss is equal to classification `difficulty' the paper's results show their method improves classification accuracy over the baselines they explored. On the other hand, \cite{li2017triple} address a similar core problem, but frame their method and it's results very differently. Their method also comprises a Generator, a Discriminator and a Classifier. However, in contrast to \cite{vandenhende2019three}, the Classifier's role is solely to predict generated images' classes. According to the authors, the Classifier and Generator then characterize the conditional distribution of images and labels (also performed by the Discriminator in many two-player GAN architectures), allowing the Discriminator to focus on identifying real and generated images. While both papers experiment with three-player GANs, the model architectures they describe amount to dual two-player games (described in \S \ref{sec:TPG}). Neither paper proposes nor analyzes the training dynamics of a three-player GAN, which will be part of the focus of this work. 

\section{Background}
\paragraph{Adversarial Game}
An adversarial game objective for two players can be represented by:
\begin{equation}
    \boldsymbol{\theta}_i^* \in \underset{\boldsymbol{\theta}_i \in \Theta_i}{\text{argmin}} \text{ } \mathcal{L}_{i} (\boldsymbol{\theta}_i, \boldsymbol{\theta}_{-i}^*)
\end{equation}

where $\theta_i$ is the parameter for player $i$, $\theta_{-i}$ represents the set of parameters for the other players, and $\mathcal{L}_{i}$ is the loss function for player $i$. A purely adversarial game is characterized by the Jacobian matrix having pure imaginary eigenvalues.

\paragraph{Nash Equilibrium}
The set of parameters that characterizes a stable state of the game in which no player can further improve their objective function locally and unilaterally.

\paragraph{Gradient Vector Field}
In games with differentiable objectives, the search for a local Nash equilibria can be achieved by taking gradient steps on the players' objectives. With the previous formulation, the gradient vector field for the three-player game is given by:
\begin{equation}
    \boldsymbol{v}(\boldsymbol{\theta}_1, \boldsymbol{\theta}_2, \boldsymbol{\theta}_3) := \left[ \nabla_{\boldsymbol{\theta}_1} \mathcal{L}_1(\boldsymbol{\theta}_1, \boldsymbol{\theta}_{-1}) \quad \nabla_{\boldsymbol{\theta}_2} \mathcal{L}_2(\boldsymbol{\theta}_2, \boldsymbol{\theta}{-2}) \quad \nabla_{\boldsymbol{\theta}_3} \mathcal{L}_3(\boldsymbol{\theta}_3, \boldsymbol{\theta}_{-3}) \right]^T
\end{equation}

and it's Jacobian is:
\begin{equation}
    \nabla\boldsymbol{v}(\boldsymbol{\theta}_1, \boldsymbol{\theta}_2, \boldsymbol{\theta}_3) := 
    \left[
    \begin{array}{ccc}
      \nabla_{\boldsymbol{\theta}_1}^2 \mathcal{L}_1(\boldsymbol{\theta}_1, \boldsymbol{\theta}_{-1})
      & \nabla_{\boldsymbol{\theta}_1}\nabla_{\boldsymbol{\theta}_2} \mathcal{L}_2(\boldsymbol{\theta}_2, \boldsymbol{\theta}_{-2})
      & \nabla_{\boldsymbol{\theta}_1}\nabla_{\boldsymbol{\theta}_3} \mathcal{L}_3(\boldsymbol{\theta}_3, \boldsymbol{\theta}_{-3}) \\
      \nabla_{\boldsymbol{\theta}_2}\nabla_{\boldsymbol{\theta}_1} \mathcal{L}_1(\boldsymbol{\theta}_1, \boldsymbol{\theta}_{-1})
      & \nabla_{\boldsymbol{\theta}_2}^2 \mathcal{L}_2(\boldsymbol{\theta}_2, \boldsymbol{\theta}_{-2})
      & \nabla_{\boldsymbol{\theta}_2}\nabla_{\boldsymbol{\theta}_3} \mathcal{L}_3(\boldsymbol{\theta}_3, \boldsymbol{\theta}_{-3}) \\
      \nabla_{\boldsymbol{\theta}_3}\nabla_{\boldsymbol{\theta}_1} \mathcal{L}_1(\boldsymbol{\theta}_1, \boldsymbol{\theta}_{-1})
      & \nabla_{\boldsymbol{\theta}_3}\nabla_{\boldsymbol{\theta}_2} \mathcal{L}_2(\boldsymbol{\theta}_2, \boldsymbol{\theta}_{-2})
      & \nabla_{\boldsymbol{\theta}_3}^2 \mathcal{L}_3(\boldsymbol{\theta}_3, \boldsymbol{\theta}_{-3}) \\
    \end{array}
    \right]    
\end{equation}

\section{Three-Player Games} \label{sec:TPG}
\subsection{Dual Bilinear Smooth Games} \label{sec:dual_bilinear}

\comment{
Considering the case where all player's update rule share the same momentum (i.e. $\beta_\theta = \beta_\phi = \beta_\psi = \beta$), the first two eigenvalues are:
\begin{equation} \label{eq:lambdas_three_players_gan}
    \lambda_{1,2} = \frac{1}{2} \left(\pm \sqrt{(\beta + 3)^2 -8} + \beta + 1\right)
\end{equation}

\todo{Can we say this without looking at the other eigenvalues?}
Note that for Eq. \ref{eq:lambdas_three_players_gan}, $\beta = -1$ turns the game into a purely adversarial game (i.e. imaginary eigenvalues).
}

\comment{
{
{    1+x,        -n,        -n,   -x, 0, 0},
{(1+x)*n, (1+y)-n*n,      -n*n, -n*x, y, 0},
{(1+x)*n,      -n*n, (1+z)-n*n, -n*x, 0, z},
{      1,         0,         0,    0, 0, 0},
{      0,         1,         0,    0, 0, 0},
{      0,         0,         1,    0, 0, 0}
}
}

\comment{
{
{    1+b,        -n,        -n,   -b, 0, 0},
{(1+b)*n, (1+b)-n*n,      -n*n, -n*b, b, 0},
{(1+b)*n,      -n*n, (1+b)-n*n, -n*b, 0, b},
{      1,         0,         0,    0, 0, 0},
{      0,         1,         0,    0, 0, 0},
{      0,         0,         1,    0, 0, 0}
}
}

\paragraph{Case: Three-Player GAN}
Even though \cite{vandenhende2019three}'s proposal consisted of three players (i.e. Generator, Discriminator and Classifier), the dynamics between them were not exhaustive: the Discriminator and Classifier don't interact directly with each other. In that sense, the game could be seen as a dual two-player game where the Generator converges to a point in between the Discriminator's distribution (i.e. producing real-looking data that is hard for Discriminator to reject) and the Classifier's distribution (i.e. producing data that is hard for the Classifier to correctly classify).

The authors define the game as:
\begin{equation}  \label{eq:three_player_gan_gan}
    \left\{
    \begin{array}{ccc}
        D^* \in \underset{D}{\text{arg max }} U_D(D, G^*)
        \\
        C^* \in \underset{C}{\text{arg min }} L_C(C, G^*)
        \\
        G^* \in \underset{G}{\text{arg min }} L_G(D^*, C^*, G)
    \end{array}
    \right.
\end{equation}
where $U_D(D, G)$ is the utility function for the Discriminator, and $L_C(C, G)$ and $L_G(D, C, G)$ are the loss functions for the Classifier and Generator, respectively.

\paragraph{Case: Triple-GAN}
Similarly, \cite{li2017triple}'s game dynamics were not exhaustive: the Generator and Classifier don't interact with each other. In that sense, the game could be seen as a dual two-player game where the Discriminator converges to a point in between the Generator's distribution and the Classifier's distribution.

The authors propose the game as:
\begin{equation}  \label{eq:triple_player_gan}
    \left\{
    \begin{array}{ccc}
        D^* \in \underset{D}{\text{arg max }} U_D(D, C^*, G^*)
        \\
        C^* \in \underset{C}{\text{arg min }} L_C(C, D^*)
        \\
        G^* \in \underset{G}{\text{arg min }} L_G(D^*, G)
    \end{array}
    \right.
\end{equation}
where $U_D(D, C, G)$ is the utility function for the Discriminator, and $L_C(C, D)$ and $L_G(D, G)$ are the loss functions for the Classifier and Generator, respectively.

\paragraph{Simplified Game Analysis}
Such dynamics can be modeled by bilinear smooth games. Following \cite{gidel2019negative}'s method, a simplified version of such a game is:
\begin{equation}  \label{eq:triple_gan}
    \underset{\boldsymbol{\theta} \in \mathbb{R}^d}{\text{max}} \ \left(
    \boldsymbol{\theta}^T\boldsymbol{c} + 
    \underset{\boldsymbol{\phi} \in \mathbb{R}^p}{\text{min}} \ \boldsymbol{\theta}^T\boldsymbol{A}\boldsymbol{\phi}  + 
    \underset{\boldsymbol{\psi} \in \mathbb{R}^q}{\text{min}} \ \boldsymbol{\theta}^T\boldsymbol{B}\boldsymbol{\psi}\right)
    \quad \text{where} \ \boldsymbol{A} \in \mathbb{R}^{d \times p}, \boldsymbol{B} \in \mathbb{R}^{d \times q}, \boldsymbol{c} \in \mathbb{R}^{d}
\end{equation}
with vector gradient field:
\begin{equation}
    \boldsymbol{v}(\boldsymbol{\theta}, \boldsymbol{\phi}, \boldsymbol{\psi}) := \left[ 
    \boldsymbol{c} + \boldsymbol{A}\boldsymbol{\phi} + \boldsymbol{B}\boldsymbol{\psi}
    \quad
    \boldsymbol{A}^T\boldsymbol{\theta}
    \quad 
    \boldsymbol{B}^T\boldsymbol{\theta}
    \right]^T
\end{equation}

Thus, the update rule is:
\begin{equation}
    \begin{array}{ccc}
      \boldsymbol{\theta}^{(t+1)} = \boldsymbol{\theta}^{(t)} + \eta 
      \left( \boldsymbol{c} + \boldsymbol{A}\boldsymbol{\phi}^{(t)} + \boldsymbol{B}\boldsymbol{\psi}^{(t)} \right)
      + \beta_\theta\left(\boldsymbol{\theta}^{(t)}-\boldsymbol{\theta}^{(t-1)}\right)
      \\
      \boldsymbol{\phi}^{(t+1)} = \boldsymbol{\phi}^{(t)} - \eta 
      \left( \boldsymbol{A}^T\boldsymbol{\theta}^{(t+1)} \right)
      + \beta_\phi\left(\boldsymbol{\phi}^{(t)}-\boldsymbol{\phi}^{(t-1)}\right)
      \\
      \boldsymbol{\psi}^{(t+1)} = \boldsymbol{\psi}^{(t)} - \eta 
      \left( \boldsymbol{B}^T\boldsymbol{\theta}^{(t+1)} \right)
      + \beta_\psi\left(\boldsymbol{\psi}^{(t)}-\boldsymbol{\psi}^{(t-1)}\right)
    \end{array}
\end{equation}

Let us rewrite the update rule with its matrix form:
\begin{equation}
    \left[
    \begin{array}{c}
        \boldsymbol{\theta}^{(t+1)}
        \\
        \boldsymbol{\phi}^{(t+1)}
        \\
        \boldsymbol{\psi}^{(t+1)}
        \\
        \boldsymbol{\theta}^{(t)}
        \\
        \boldsymbol{\phi}^{(t)}
        \\
        \boldsymbol{\psi}^{(t)}
        \\
    \end{array}
    \right]
    := \boldsymbol{F}_{\eta,\beta}
    \left[
    \begin{array}{c}
        \boldsymbol{\theta}^{(t)}
        \\
        \boldsymbol{\phi}^{(t)}
        \\
        \boldsymbol{\psi}^{(t)}
        \\
        \boldsymbol{\theta}^{(t-1)}
        \\
        \boldsymbol{\phi}^{(t-1)}
        \\
        \boldsymbol{\psi}^{(t-1)}
        \\
    \end{array}
    \right]
    + \boldsymbol{g}_\eta \cdot \boldsymbol{c}
\end{equation}

Hence, the matrix $\boldsymbol{F}_{\eta,\beta}$ is:
\begin{equation} \label{eq:f_eta_beta2}
    \boldsymbol{F}_{\eta,\beta}= 
    \left[
    \begin{array}{cccccc}
      (1+\beta_\theta)\boldsymbol{I}_d
      & \eta\boldsymbol{A}
      & \eta\boldsymbol{B}
      & -\beta_\theta\boldsymbol{I}_{d}
      & \boldsymbol{0}_{d p}
      & \boldsymbol{0}_{d q}
      \\
      -(1+\beta_\theta)\eta\boldsymbol{A}^T
      & (1+\beta_\phi)\boldsymbol{I}_p -\eta^2\boldsymbol{A}^T\boldsymbol{A}
      & -\eta^2\boldsymbol{A}^T\boldsymbol{B}
      & \eta\beta_\theta\boldsymbol{A}^T
      & \beta_\phi\boldsymbol{I}_p
      & \boldsymbol{0}_{p q}
      \\
      -(1+\beta_\theta)\eta\boldsymbol{B}^T
      & -\eta^2\boldsymbol{B}^T\boldsymbol{A}
      & (1+\beta_\psi)\boldsymbol{I}_q -\eta^2\boldsymbol{B}^T\boldsymbol{B}
      & \eta\beta_\theta\boldsymbol{B}^T
      & \boldsymbol{0}_{q p}
      & \beta_\psi\boldsymbol{I}_q
      \\
      \boldsymbol{I}_d
      & \boldsymbol{0}_{d p}
      & \boldsymbol{0}_{d q}
      & \boldsymbol{0}_{d d}
      & \boldsymbol{0}_{d p}
      & \boldsymbol{0}_{d q}
      \\
      \boldsymbol{0}_{p d}
      & \boldsymbol{I}_p
      & \boldsymbol{0}_{p q}
      & \boldsymbol{0}_{p d}
      & \boldsymbol{0}_{p p}
      & \boldsymbol{0}_{p q}
      \\
      \boldsymbol{0}_{q d}
      & \boldsymbol{0}_{q p}
      & \boldsymbol{I}_q
      & \boldsymbol{0}_{q d}
      & \boldsymbol{0}_{q p}
      & \boldsymbol{0}_{q q}
    \end{array}
    \right]    
\end{equation}

and:
\begin{equation}
    \boldsymbol{g}_\eta = 
    \left[
    \begin{array}{c}
        \eta\boldsymbol{I}^d
        \\
        -\eta^2\boldsymbol{A}^T
        \\
        -\eta^2\boldsymbol{B}^T
        \\
        \boldsymbol{0}_{d d}
        \\
        \boldsymbol{0}_{p d}
        \\
        \boldsymbol{0}_{q d}
    \end{array}
    \right]
\end{equation}

For the special case where $d=p=q=1$ and $\boldsymbol{A} = \boldsymbol{B} = 1$ The eigenvalues of such $\boldsymbol{F}_{\eta,\beta}$ are the roots of:
\begin{dmath} \label{eq:polychar_three_player_triple_gan}
    \lambda^6
    - (\beta_\theta + \beta_\phi + \beta_\psi - 2n^2 + 3)\lambda^5
    + (\beta_\theta \beta_\phi + \beta_\theta \beta_\psi + \beta_\phi \beta_\psi + 3\beta_\theta  - (n^2 -1)(\beta_\phi + \beta_\psi) - 2n^2 + 3)\lambda^4
    - (\beta_\theta \beta_\phi \beta_\psi + \beta_\theta \beta_\phi + \beta_\theta \beta_\psi - \beta_\phi \beta_\psi + 3\beta_\theta + (n^2 -1)(\beta_\phi + \beta_\psi) + 1)\lambda^3
    - (\beta_\theta \beta_\phi \beta_\psi + \beta_\theta \beta_\phi + \beta_\theta \beta_\psi 
     + \beta_\phi \beta_\psi - \beta_\theta + \beta_\phi + \beta_\psi)\lambda^2
    + (\beta_\theta \beta_\phi \beta_\psi  + \beta_\theta \beta_\phi + \beta_\theta \beta_\psi - \beta_\phi \beta_\psi)\lambda
    + \beta_\theta \beta_\phi \beta_\psi = 0
\end{dmath}

\comment{
{
{    1+x,         n,         n,   -x, 0, 0},
{-(1+x)*n, (1+y)-n*n,      -n*n, n*x, y, 0},
{-(1+x)*n,      -n*n, (1+z)-n*n, n*x, 0, z},
{      1,         0,         0,    0, 0, 0},
{      0,         1,         0,    0, 0, 0},
{      0,         0,         1,    0, 0, 0}
}
}

\comment{
{
{    1+b,         n,         n,   -b, 0, 0},
{-(1+b)*n, (1+b)-n*n,      -n*n, n*b, b, 0},
{-(1+b)*n,      -n*n, (1+b)-n*n, n*b, 0, b},
{      1,         0,         0,    0, 0, 0},
{      0,         1,         0,    0, 0, 0},
{      0,         0,         1,    0, 0, 0}
}
}

\subsection{Trilinear Smooth Game} \label{sec:trilinear_game}
Following the definition provided in \cite{gidel2019negative} for simplified bilinear smooth games, we propose a simplified trlinear smooth game, a purely adversarial game described by:
\begin{equation}  \label{eq:our_game}
    \underset{\boldsymbol{\theta} \in \mathbb{R}^d}{\text{max}} \
    \underset{\boldsymbol{\phi} \in \mathbb{R}^p}{\text{min}} \ \underset{\boldsymbol{\psi} \in \mathbb{R}^q}{\text{min}} \boldsymbol{\theta}^T\boldsymbol{A}\boldsymbol{\phi} \boldsymbol{b}^T\boldsymbol{\psi}, \quad \text{where} \ \boldsymbol{A} \in \mathbb{R}^{d \times p} \text{, } \ \boldsymbol{b} \in \mathbb{R}^{q}
\end{equation}

with vector gradient field:
\begin{equation}
    \boldsymbol{v}(\boldsymbol{\theta}, \boldsymbol{\phi}, \boldsymbol{\psi}) := \left[ 
    \boldsymbol{A}\boldsymbol{\phi} \boldsymbol{b}^T\boldsymbol{\psi}
    \quad
    \boldsymbol{A}^T\boldsymbol{\theta}\boldsymbol{\psi}^T \boldsymbol{b}
    \quad 
    \boldsymbol{b}\boldsymbol{\phi}^T \boldsymbol{A}^T\boldsymbol{\theta}
    \right]^T
\end{equation}

Thus, the update rule\footnote{Note that given the non-linear nature of the update rule, it is not trivial to rewrite it as a linear transformation similar to Eq. \ref{eq:f_eta_beta2}.} is:
\begin{equation} \label{eq:non_linear}
    \begin{array}{ccc}
    \boldsymbol{\theta}^{(t+1)} = \boldsymbol{\theta}^{(t)} + \eta 
      \left( \boldsymbol{A}\boldsymbol{\phi}^{(t)} \boldsymbol{b}^T\boldsymbol{\psi}^{(t)} \right)
      + \beta\left(\boldsymbol{\theta}^{(t)}-\boldsymbol{\theta}^{(t-1)}\right)
      \\
      \boldsymbol{\phi}^{(t+1)} = \boldsymbol{\phi}^{(t)} - \eta 
      \left( \boldsymbol{A}^T\boldsymbol{\theta}^{(t+1)}\left(\boldsymbol{\psi}^{(t)}\right)^T \boldsymbol{b} \right)
      + \beta\left(\boldsymbol{\phi}^{(t)}-\boldsymbol{\phi}^{(t-1)}\right)
      \\
      \boldsymbol{\psi}^{(t+1)} = \boldsymbol{\psi}^{(t)} - \eta 
      \left( \boldsymbol{b}\left(\boldsymbol{\phi}^{(t+1)}\right)^T \boldsymbol{A}^T\boldsymbol{\theta}^{(t+1)} \right)
      + \beta\left(\boldsymbol{\psi}^{(t)}-\boldsymbol{\psi}^{(t-1)}\right)
    \end{array}
\end{equation}

\subsection{SquidGAN}  \label{sec:squid_gan}
Following the trilinear smooth game presented above, we propose SquidGAN: an adversarial game where each network interacts with the other two players. For this setting we have a real set of data-label pairs ${(\boldsymbol{x}, y) \sim p_{\text{data}}(\boldsymbol{x}, y)}$, as well as input noise variables $p_{\boldsymbol{x}_z}(\boldsymbol{x}_z)$ and randomly selected labels $y_z \sim \mathcal{U}\{y\}_\neq$\footnote{$\mathcal{U}\{y\}_\neq$ represents a discrete uniform random distribution over the set of unique values of $y$.}. We denote the latter pairs by $(\boldsymbol{x}_z, y_z) \sim p_z(\boldsymbol{x}_z, y_z)$.

\begin{figure}[t]
    \centering
    \includegraphics[width=\textwidth]{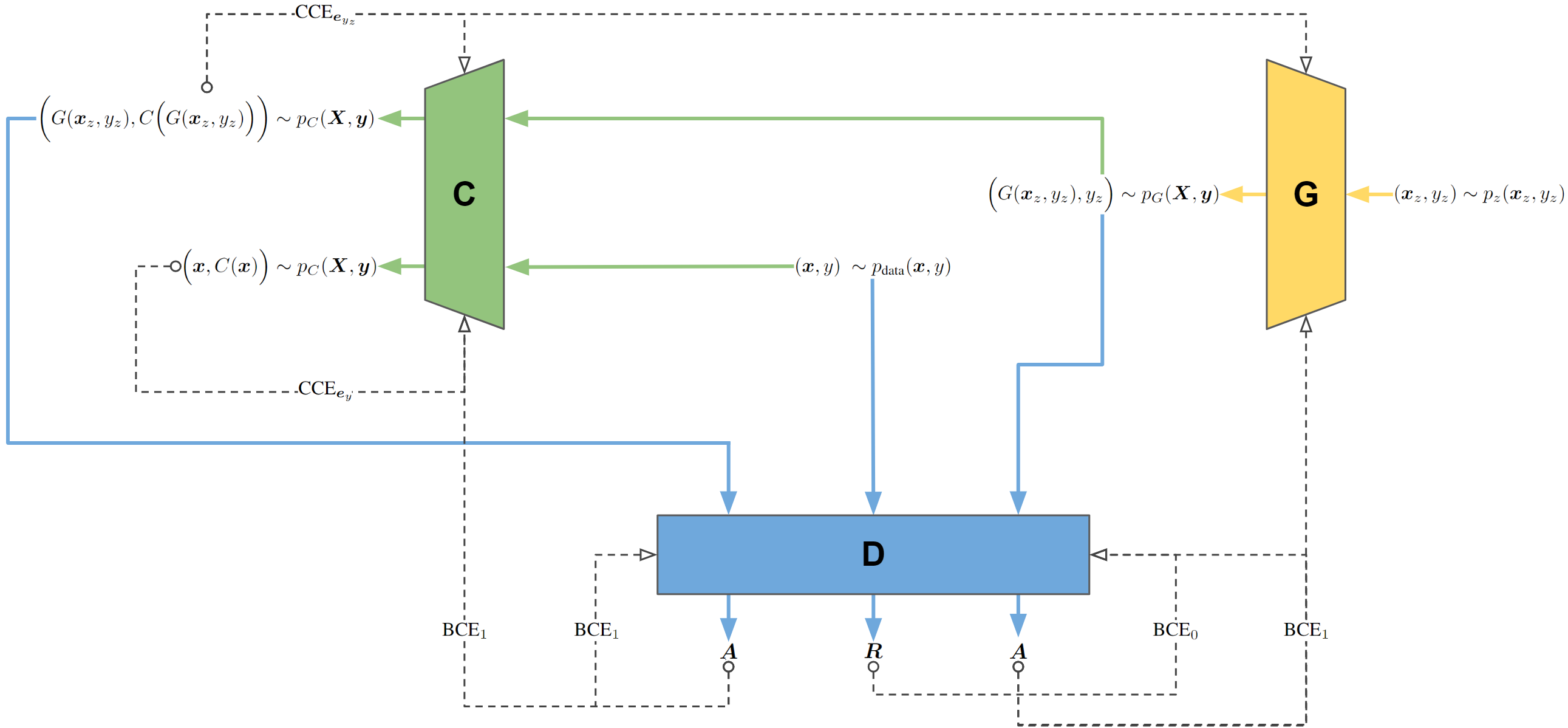}
    \caption{Illustration of SquidGAN. The colored solid lines represent the data (ordered pairs) dependencies between networks, including the true distribution. The black dashed lines represent the back-propagation (loss) dependency between the networks. $\text{BCE}_i$ is the binary cross-entropy loss with respect to $i$, and $\text{CCE}_{\boldsymbol{i}}$ is the categorical cross-entropy loss with respect to $\boldsymbol{i}$. `A' denotes acceptance, while `R' denotes rejection, and represent each network's interest.}
    \label{fig:gan_illustration}
\end{figure}

The three models are: (a) a Classifier $C$ that aims to maximize the probability of correctly classifying real data as well as the generated samples, while at the same time maximizing the probability that $D$ doesn't reject the pair given by $G$'s output and its own classification; (b) a Discriminator $D$ that maximizes the probability of assigning the correct true/false label to the real examples and the samples from $G$; and (c) a Generator $G$ that maximizes the probability that $D$ classifies its sample as real, as well as the probability of generating samples that $C$ classifies correctly (see Figure \ref{fig:gan_illustration} for the illustration of SquidGAN).

The GAN's game is defined as:
\begin{equation}  \label{eq:our_game_gan}
    \left\{
    \begin{array}{ccc}
        D^* \in \underset{D}{\text{arg min }} L_D(D, C^*, G^*)
        \\
        C^* \in \underset{C}{\text{arg max }} U_C(D^*, C, G^*)
        \\
        G^* \in \underset{G}{\text{arg max }} U_G(D^*, C^*, G)
    \end{array}
    \right.
\end{equation}

\break
where

\begin{align}
    \begin{split}
        L_D(D, C, G) &\triangleq
        \mathbb{E}_{(\boldsymbol{x}, y) \sim p_{\text{data}}(\boldsymbol{x}, y)}\Big[\text{BCE}\Big(D(\boldsymbol{x}, y), 0\Big)\Big] \\
        & + \mathbb{E}_{(\boldsymbol{x}_z, y_z) \sim p_z(\boldsymbol{x}_z, y_z)}\Bigg[\text{BCE}\Bigg(D\bigg(G(\boldsymbol{x}_z, y_z), C\Big(G(\boldsymbol{x}_z, y_z)\Big)\bigg), 1\Bigg)\Bigg]
    \end{split}
\end{align}

\begin{align}
    \begin{split}
        U_C(D, C, G) &\triangleq
        \mathbb{E}_{(\boldsymbol{x}, y) \sim p_{\text{data}}(\boldsymbol{x}, y)}\Big[\text{CCE}\Big(C(\boldsymbol{x}), \boldsymbol{e}_y\Big)\Big] \\
        & + \mathbb{E}_{(\boldsymbol{x}_z, y_z) \sim p_z(\boldsymbol{x}_z, y_z)}\bigg[\text{CCE}\bigg(C\Big(G(\boldsymbol{x}_z, y_z)\Big), \boldsymbol{e}_{y_z}\bigg) \\
        & \qquad + \text{BCE}\Bigg(D\bigg(G(\boldsymbol{x}_z, y_z), C\Big(G(\boldsymbol{x}_z, y_z)\Big)\bigg), 1\Bigg)\Bigg]
    \end{split}
\end{align}

\begin{align}
    \begin{split}
        U_G(D, C, G) &\triangleq
        \mathbb{E}_{(\boldsymbol{x}_z, y_z) \sim p_z(\boldsymbol{x}_z, y_z)}\bigg[\text{BCE}\bigg(D\Big(G(\boldsymbol{x}_z, y_z), y_z\Big), 1\bigg) + \text{CCE}\bigg(C\Big(G(\boldsymbol{x}_z, y_z)\Big), \boldsymbol{e}_{y_z}\bigg)\bigg]
    \end{split}
\end{align}

\section{Experiments and Discussion}
\comment{
\todo{Include at least one of:
    \begin{itemize}
        \item An experimental comparison of the results of your method compared with a baseline. Qualitative evaluation is OK.
        \item An experiment demonstrating a property that your model has that a baseline model does not. Experiments should also include a description of how you prepared your datasets, how you trained your model, and any tricks you used to get it to work.
        \item An experiment that reveals interesting properties of or relationships between existing methods.
        \item A proof of a theorem or conjecture, or an interesting counterexample
    \end{itemize}
Toy data is OK! The point is to help the reader understand why or when we would want to use one approach over another, or to understand something better. Try to summarize the main takeaways. Negative results are fine, as long as you have an insightful and well supported explanation
}
}

\begin{figure}[t]
    \centering
    \includegraphics[width=\textwidth]{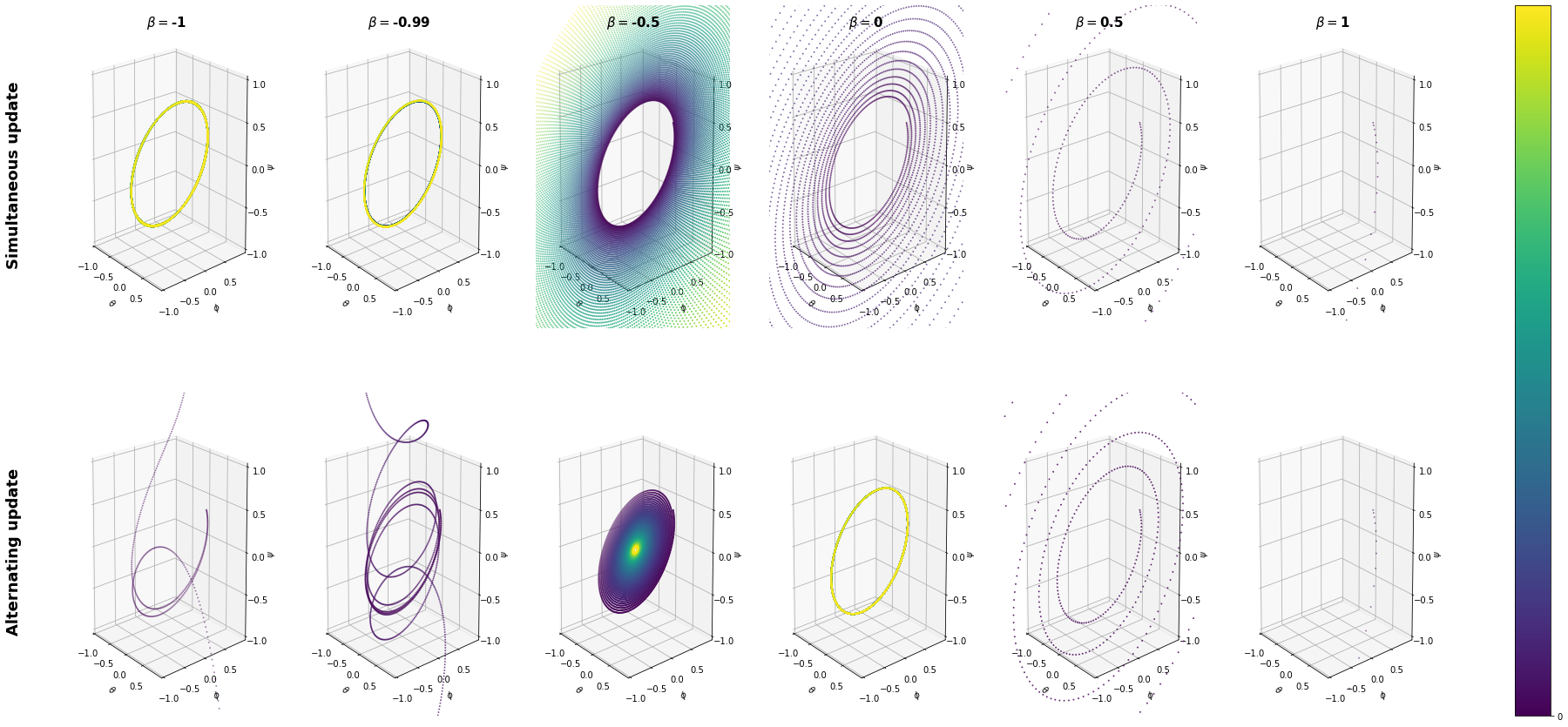}
    \caption{The effect of gradient steps' order and in a dual two-player max-min bilinear game. The color represents the time elapsed in the game. (Step size of $2$e-$2$, and $25$k iterations.)}
    \label{fig:triple_gan}
\end{figure}

\paragraph{Dual Bilinear Smooth Game (Cases: Three-Player GAN and Triple-GAN)}
[Fig. \ref{fig:triple_gan}] We investigate the game dynamics of the optimization setup given by Eq. \ref{eq:triple_gan} with $\boldsymbol{\theta}, \boldsymbol{\phi}, \boldsymbol{\psi} \in \mathbb{R}$, $\boldsymbol{A} = \boldsymbol{B} = 1$ and $\beta_\theta = \beta_\phi = \beta_\psi = \beta$, examining different update orders (i.e. alternating versus simultaneous gradient steps, and multiple momentum values). As mentioned above, this game can be characterized as the sum of two two-player games. The result is a game played on a two-dimensional plane projected into three-dimensional space—even for $\beta=-1$ and $\beta=-0.99$ for the alternating update in Figure \ref{fig:triple_gan}. Interestingly, the dynamics of that game are very similar to the two-player bilinear smooth game explored by \cite{gidel2019negative}. The game only reaches equilibrium when alternating update order is applied with negative momentum.


\begin{figure}[t]
    \centering
    \includegraphics[width=\textwidth]{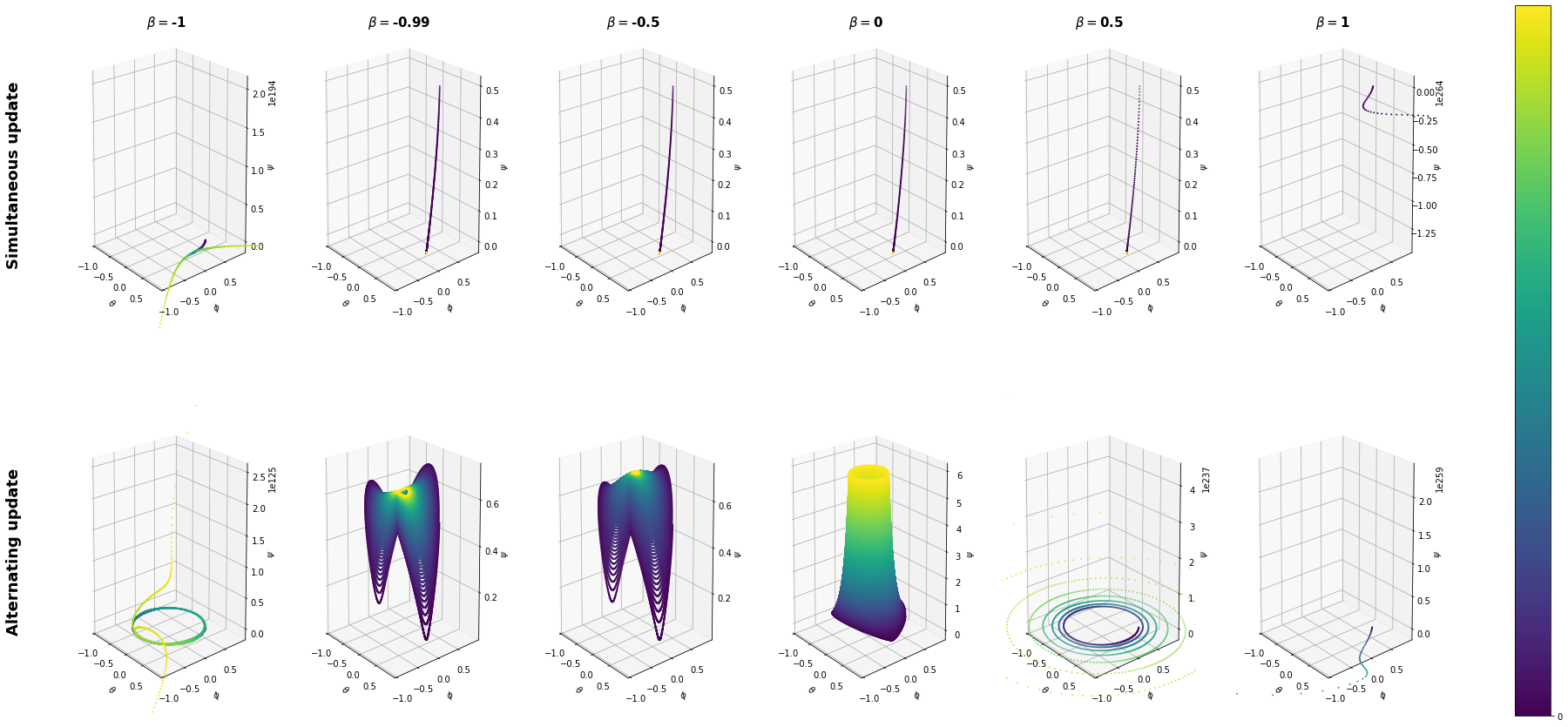}
    \caption{The effect of gradient steps' order and in a trilinear max-min-min game. The color represents the time elapsed in the game. (Step size of $2$e-$2$, and $100$k iterations.)}
    \label{fig:our_game}
\end{figure}

\paragraph{Trilinear Smooth Game (Case: SquidGAN)}
[Fig. \ref{fig:our_game}] We show the game dynamics of the optimization setup given by Eq. \ref{eq:our_game} with $\boldsymbol{\theta}, \boldsymbol{\phi}, \boldsymbol{\psi} \in \mathbb{R}$,  $\boldsymbol{A} = \boldsymbol{b} = 1$ and $\beta_\theta = \beta_\phi = \beta_\psi$, from different update rules (i.e. alternating versus simultaneous gradient steps, and multiple momentum values). In contrast to games exploring Eq. \ref{eq:triple_gan}, the update rule for each player in a simplified trilinear game includes terms representing both other players. Consequently, the training dynamics observed are markedly different. Most notably, these interactions result in a 3-dimensional optimization path. Unlike either two-player or dual two-player games, simplified trilinear games with equal momentum values for all players do not reach Nash equilibrium.

In most cases two players reach their optimal value and one player does not. However, the third player does reach a stable point. This result can be seen in both the simultaneous and alternating conditions paired with a negative momentum (value below $-1$) displayed in Figure \ref{fig:our_game}. Note that the closer the momentum value is to $-1$, the closer the game's equilibrium is to the Nash equilibrium. Interestingly, the simultaneous case also reaches an equilibrium similar to that yielded by alternating updates regardless of whether momentum is positive or negative (in fewer steps, in fact). The player that reaches sub-optimal equilibrium is defined by the order of updates in the alternating case and is the maximizing player in the simultaneous update case.

\begin{figure}[t]
    \centering
    \includegraphics[scale=0.45]{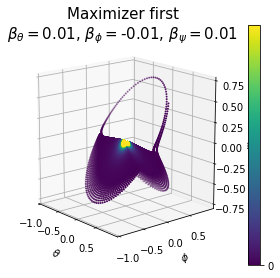}
    \caption{Optimization path for a specific set of $\beta$ converging to the Nash equilibrium of the trilinear smooth game. The color represents the time elapsed in the game. (Step size of $1$e-$1$, and $500$k iterations.)}
    \label{fig:nash_eq}
\end{figure}

The simplified trilinear game presents the opportunity for a new update order designed to ameliorate the sub-optimal equilibrium produced by alternating updates. This new update rule updates the maximizer first then the minimizers simultaneously. Our experiments show that while this update rule behaves similar to simultaneous updates when all three players' momentum values are equal, but creates very different optimization paths when they are not. For example, setting the maximizer's momentum value to less than zero allows for convergence given a variety of minimizer momentum values, both positive and negative as shown in Figure \ref{fig:nash_eq}.

\begin{figure}
    \subfigure[Simultaneous updates]{\includegraphics[width=0.32\textwidth]{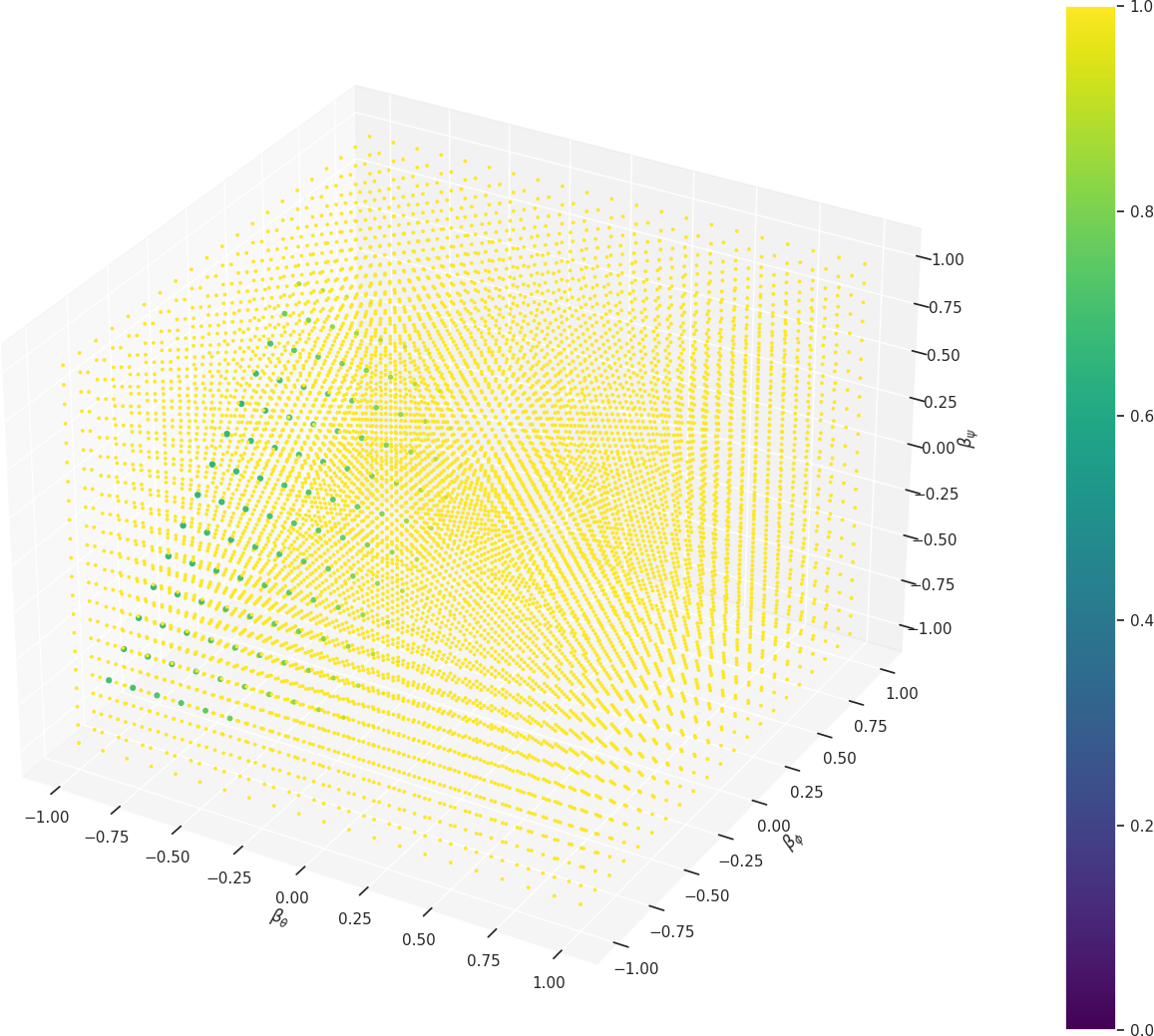}}
    \subfigure[Alternating updates]{\includegraphics[width=0.32\textwidth]{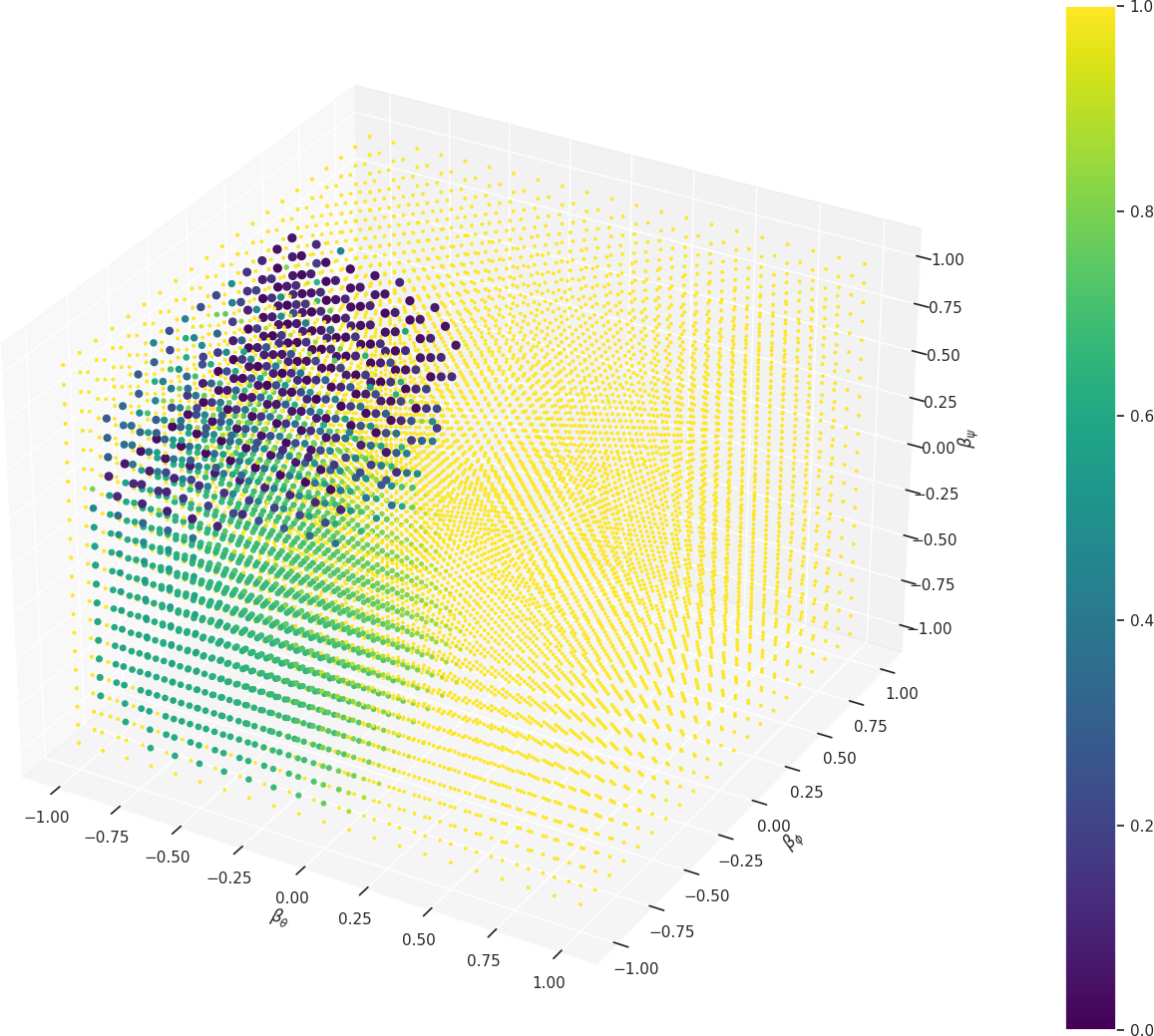}}
    \subfigure[Maximizer-first updates]{\includegraphics[width=0.32\textwidth]{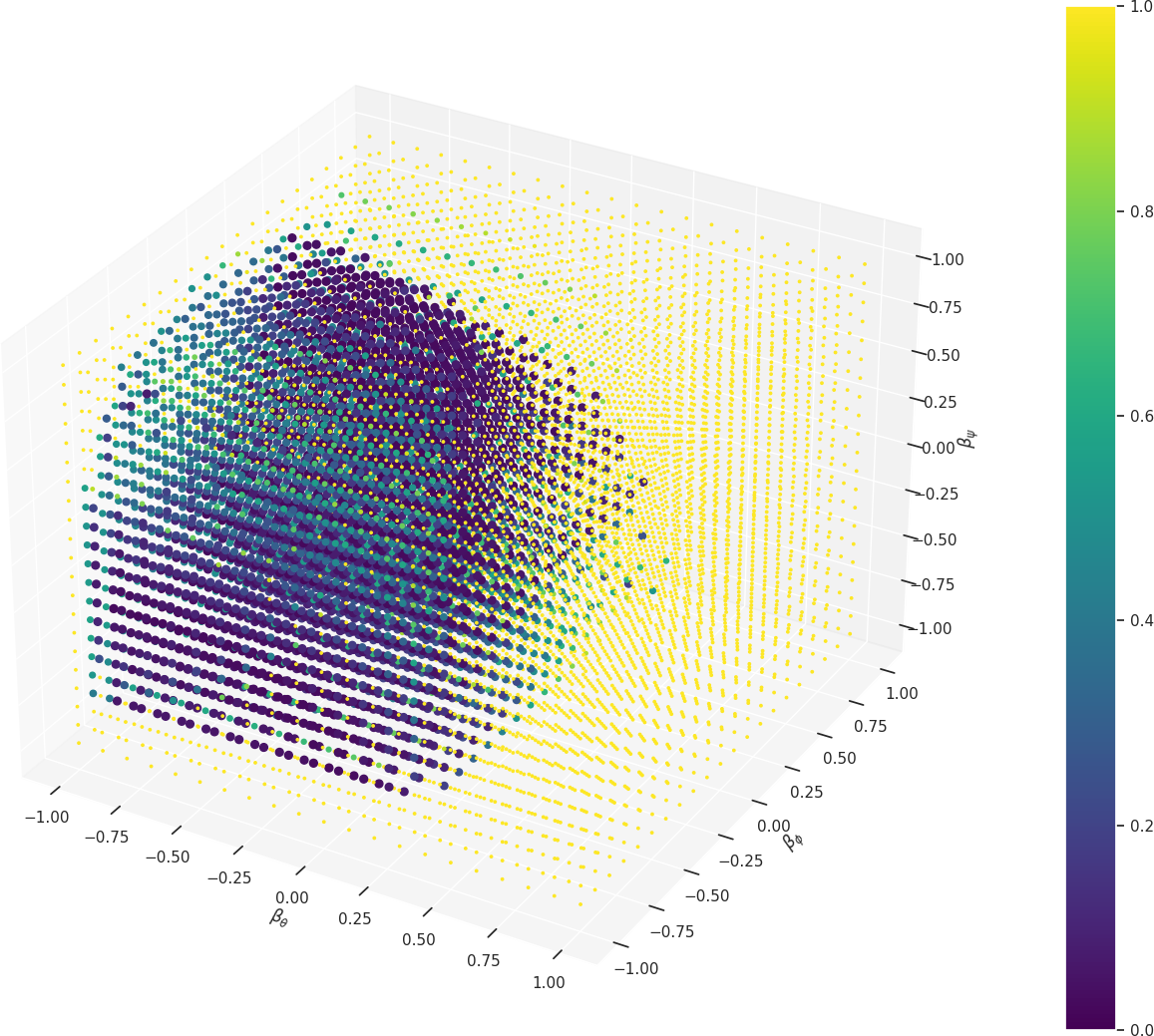}}
    \caption{A grid search over momentum values for all three players in a simplified trilinear game calculated with a step size of $2$e-$2$ after $100$k iterations. The size of bubble and color represent the resulting equilibrium point's euclidean distance to the Nash equilibrium ($\boldsymbol{0}$). We performed the experiment using different update rules.}
    \label{fig:grid_search}
\end{figure}

In order to more fully explore the interactions between players with different momentum values for every update order conditions, we performed a grid search\footnote{See \S \ref{appendix:interesting_cases} for a subset of interesting cases.}. The grid search we implemented explored $20$ momentum values between $-1$ and $1$ for each player and condition, totaling $24,000$ vertices. Each game was played for $100,000$ iterations. The euclidean distance between game's endpoint and $\boldsymbol{0}$ was calculated, capped at $1$ in order to make our plots readable. The results are displayed for the simultaneous, alternating, and maximizer-first conditions in Figure \ref{fig:grid_search}.

\begin{figure}[t]
\centering
    \subfigure[SquidGAN]{\includegraphics[width=0.43\textwidth]{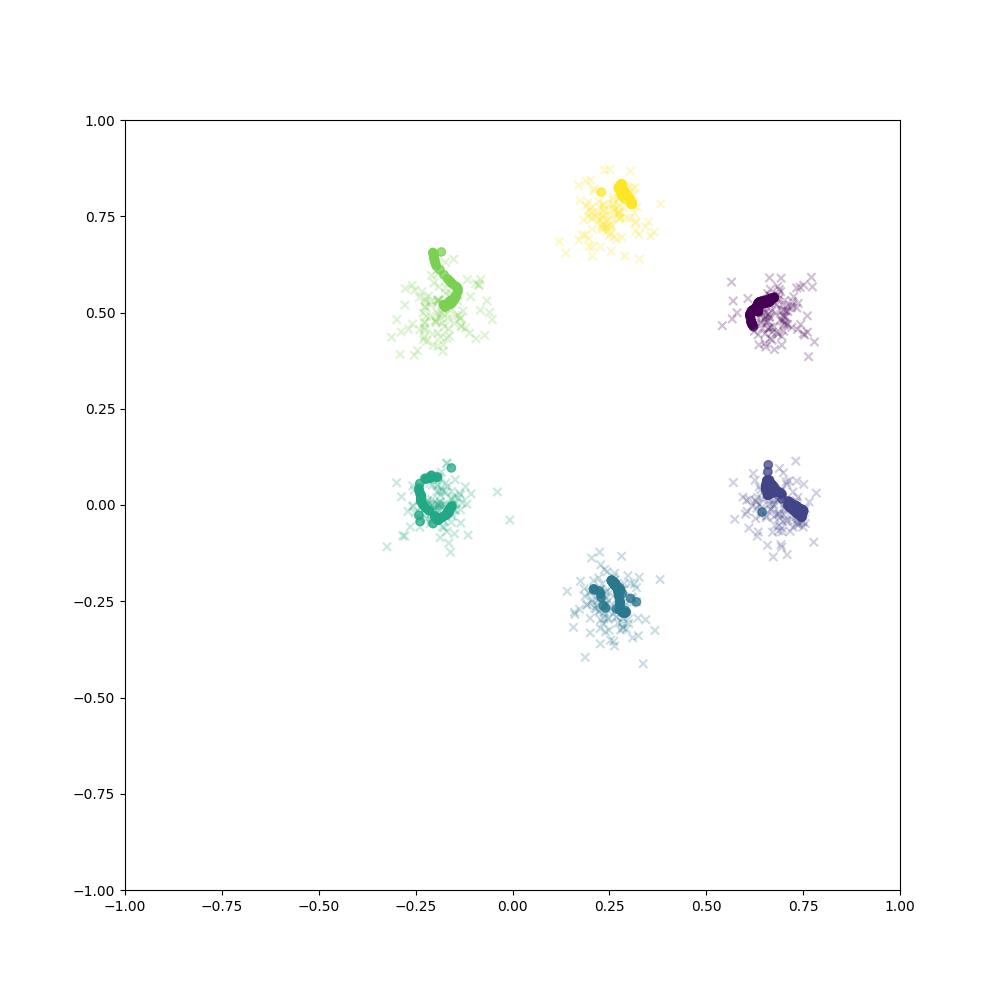}}
    \subfigure[Triple-GAN]{\includegraphics[width=0.43\textwidth]{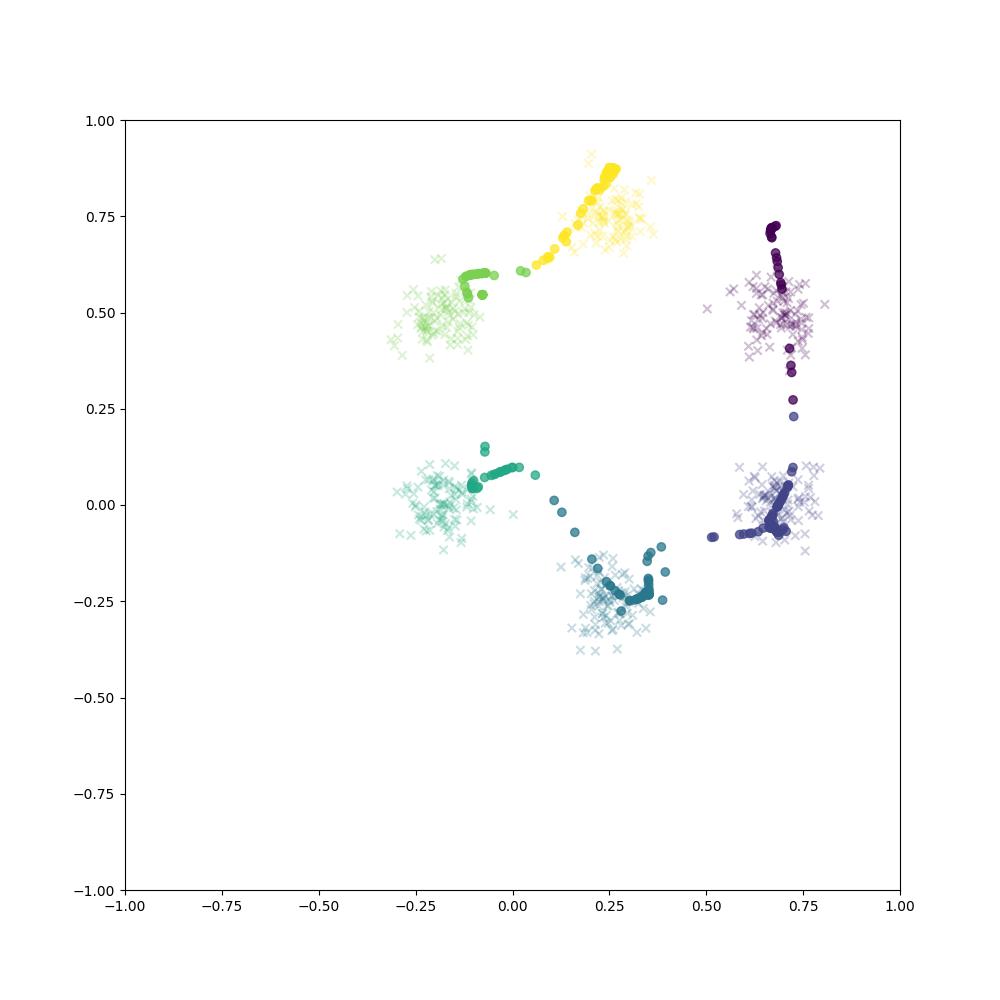}}
    \caption{Outputs after $180$k iterations of training. These architectures were designed so that the Generator would output a point in one of six Gaussian distributions. The faint colors in the background of each image represent the Gaussian and their assigned classes while the brighter dots represent the output of the Generator and the class it was asked to generate.}
    \label{fig:gaussians}
\end{figure}

\paragraph{SquidGAN}
We apply the SquidGAN framework (described in \S \ref{sec:squid_gan}) to a mixture of six Gaussian distributions as show in Figure \ref{fig:gaussians}. Each Gaussian distribution is assigned a class and that class is provided to the Generator as an input. The Generator was trained alongside a Discriminator and Classifier as described in Eq. \ref{eq:our_game_gan}. Interestingly, SquidGAN generates much tighter clusters than Triple-GAN, although we can make no claims about comparative performance without additional experiments.

\section{Limitations \& Future Work}

Our work is exploratory and limited in a number of ways. First, both the difficulty of analytically describing a three-player game with distinct momentums for each player and the non-linear nature of the dynamics when each player interacts with the others, limited the depth of our analysis. While we provided experimental results, we could not provide proofs to support them. Future work could extend our analysis. Second, we only explored one, illustrative, application of a three-player GAN. This clearly limits the scope of this work's claims. Future work could apply three-player GANs to other problems. Third, we make no claims about the utility of the SquidGAN vis-a-vis other GAN architectures. This follows from the second limitation and the scope of this paper. Applying SquidGAN or a similar architecture to datasets commonly used to compare GAN architectures like the MNIST or CIFAR family of dataset is a straightforward extension of this work.

The addition of new players with distinct objectives provides an interesting tool for future researchers. Such players can represent arbitrary interests not present in the a typical loss function or optimizer. In SquidGAN, we used a new player to force the Generator and Discriminator to be sensitive to classes. Extending that method to new applications could lead to interesting results.

\section{Conclusions}

This work analyzed the training dynamics of three-player games which, in contrast to other work that examine dual two-player games, include explicit interactions between all players. We analytically and empirically described the differences between dual two-player and three-player games. Further, we experimented with a trilinear smooth game showing under which conditions it reaches equilibrium. 

Our analysis of dual bilinear games showed that adding a third player that doesn't interact with the other two, corresponds to a rotation of one of the bilinear games. Even more, we showed how the simultaneous vs. alternating update rules, as well as different momentums, affect the convergence of such games.

In our experimental examination of trilinear smooth games we found that while trilinear games reach an equilibrium, they do not reach Nash equilibrium under almost all conditions. This is the result of direct competition across three updates which include information from all players. However, it is worth noting that in many cases, two of the three players reach their objective (i.e., 0 for both). The player who does not reach its objective is determined by the order of updates or is the maximizing player. 

True three-player games also allow for a type of update not explored in other literature (i.e. maximizer-first update) which calls for first updating the maximizing agent, then updating both minimizing agents simultaneously\footnote{Respectively, the minimizer-first update rule for min-max-max games.}. We found that maximizer-first updates allow for the game to reach an equilibrium, often closer to the Nash equilibrium, under more hyper-parameter combinations than other types of updates.

Finally, we implemented a three-player GAN, called SquidGAN. We applied SquidGAN to a mixture of Gaussian distributions in such a way that it would learn to produce points within a specified distribution. Our experiment validates that a three-player GAN can produce a working model. This result adds to other similar papers have tested dual two-player games.

\comment{
\section{List of contributions}
The authors collaborated on every aspect of this work. The contributions listed below represent the components each team member lead.
\begin{itemize}
    \item \textbf{Kenneth Christofferson}:
    \begin{itemize}
        \item Experiments and code
    \end{itemize}
    \item \textbf{Fernando Yanez}:
    \begin{itemize}
        \item Theoretical analysis and code
    \end{itemize}
\end{itemize}
}

\bibliographystyle{plainnat}
\bibliography{references}

\comment{
assuming that an equilibrium exists and using the translations $\boldsymbol{\theta}_1 \rightarrow \boldsymbol{\theta}_1 - \boldsymbol{d'} - \boldsymbol{e'}$, $\boldsymbol{\theta}_2 \rightarrow \boldsymbol{\theta}_2 - \boldsymbol{c}'$, and $\boldsymbol{\theta}_3 \rightarrow \boldsymbol{\theta}_3 - \boldsymbol{c}'$ where $\boldsymbol{c} = \boldsymbol{A}\boldsymbol{c'}$, $\boldsymbol{d} = \boldsymbol{A}\boldsymbol{d'}$ and $\boldsymbol{e} = \boldsymbol{A}\boldsymbol{e'}$, we can assume WLOG that $p \geq d$, $q \geq d$, $\boldsymbol{c} = \boldsymbol{d} = \boldsymbol{e} = \boldsymbol{0}$ (See proof in Appendix \ref{appendix:smoothgame}.)
}

\appendix

\break

\section*{Appendices}
\addcontentsline{toc}{section}{Appendices}
\renewcommand{\thesubsection}{\Alph{subsection}}

\subsection{Results for different momentums per player}
\label{appendix:interesting_cases}

\begin{figure}[h]
    \centering
    \includegraphics[scale=0.31]{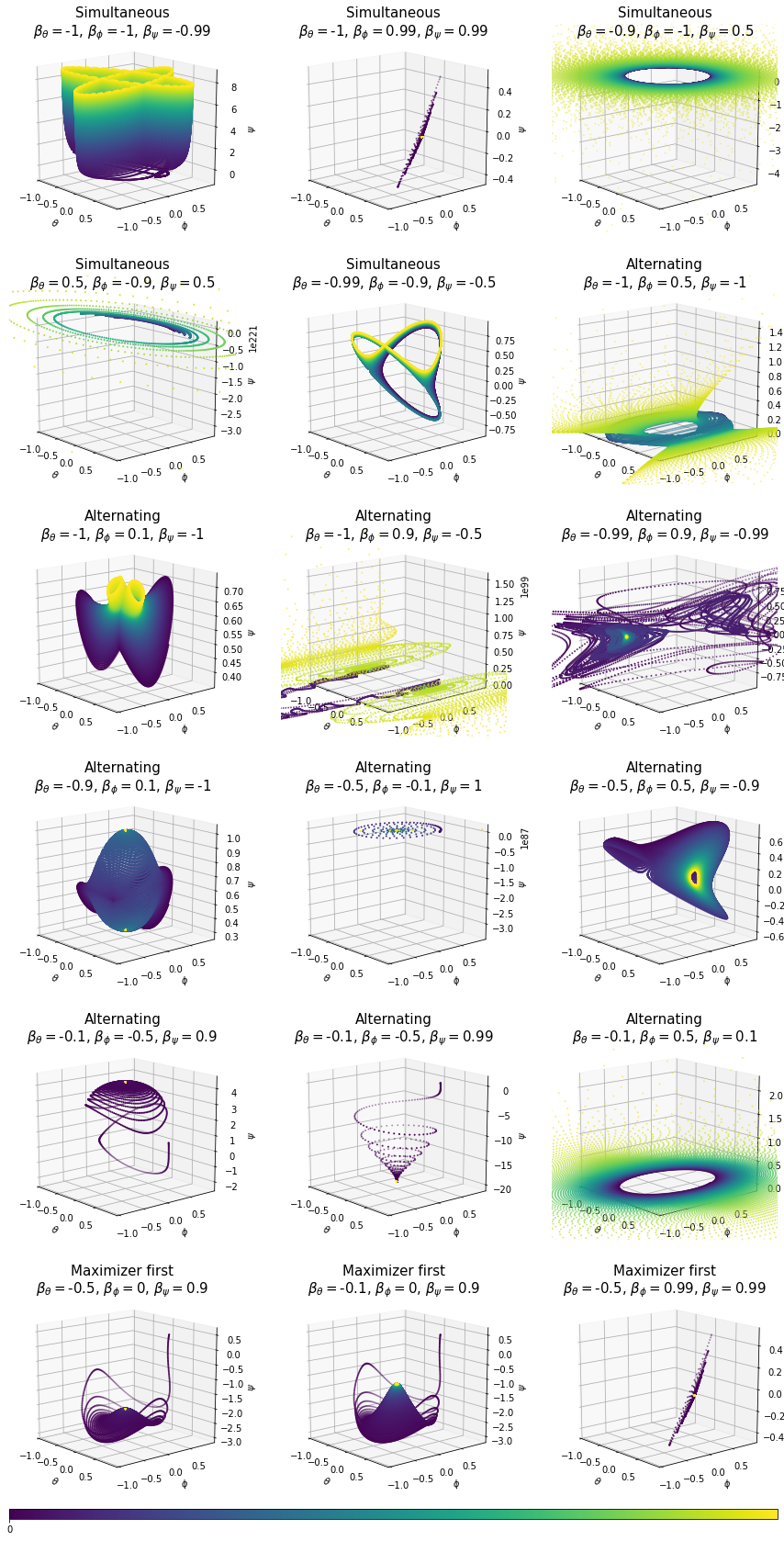}
    \caption{Optimization paths in a trilinear smooth game. The color represents the time elapsed in the game. The title of each plot shows the update rule setting, as well as the different momentums. (Step size of $2$e-$2$, and $500$k iterations.)}
    \label{fig:interesting_cases}
\end{figure}

\end{document}